\documentclass[10pt,twocolumn,letterpaper]{article}

\usepackage{iccv}
\usepackage{times}
\usepackage{epsfig}
\usepackage{graphicx}
\usepackage{amsmath}
\usepackage{amssymb}

\usepackage{multirow}
\usepackage{makecell}
\usepackage{subfigure}
\usepackage{amsfonts}
\usepackage{bm}
\usepackage{booktabs}
\usepackage{bbding}
\usepackage{amsmath}
\usepackage{pdfpages}

\usepackage[breaklinks=true,bookmarks=false]{hyperref}

\iccvfinalcopy 

\def\iccvPaperID{****} 

\ificcvfinal\fi

\begin{document}

\title{Exploring Classification Equilibrium in Long-Tailed Object Detection}

\author{Chengjian Feng\\
Intellifusion Inc.\\
{\tt\small feng.chengjian@intellif.com}
\and
Yujie Zhong$^{*}$\\
Meituan Inc.\\
{\tt\small zhongyujie@meituan.com}
\and
Weilin huang\\
Tao Technology, Alibaba Group \\
{\tt\small weilin.hwl@alibaba-inc.com}
}

\maketitle
\ificcvfinal\thispagestyle{empty}\fi

\begin{abstract}
The conventional detectors tend to make imbalanced classification and suffer performance drop, when the distribution of the training data is severely skewed. In this paper, we propose to use the mean classification score to indicate the classification accuracy for each category during training. Based on this indicator, we balance the classification via an Equilibrium Loss~(EBL) and a Memory-augmented Feature Sampling~(MFS) method. Specifically, EBL increases the intensity of the adjustment of the decision boundary for the weak classes by a designed score-guided loss margin between any two classes. On the other hand, MFS improves the frequency and accuracy of the adjustment of the decision boundary for the weak classes through over-sampling the instance features of those classes. Therefore, EBL and MFS work collaboratively for finding the classification equilibrium in long-tailed detection, and dramatically improve the performance of tail classes while maintaining or even improving the performance of head classes. We conduct experiments on LVIS using Mask R-CNN with various backbones including ResNet-50-FPN and ResNet-101-FPN to show the superiority of the proposed method. It improves the detection performance of tail classes by 15.6 AP, and outperforms the most recent long-tailed object detectors by more than 1 AP. Code is available at~\url{https://github.com/fcjian/LOCE}.
\vspace{-5mm}
\end{abstract}

\begin{figure}[ht]
		\centering
		\includegraphics[height=3cm,trim=0 0 0 35,clip]{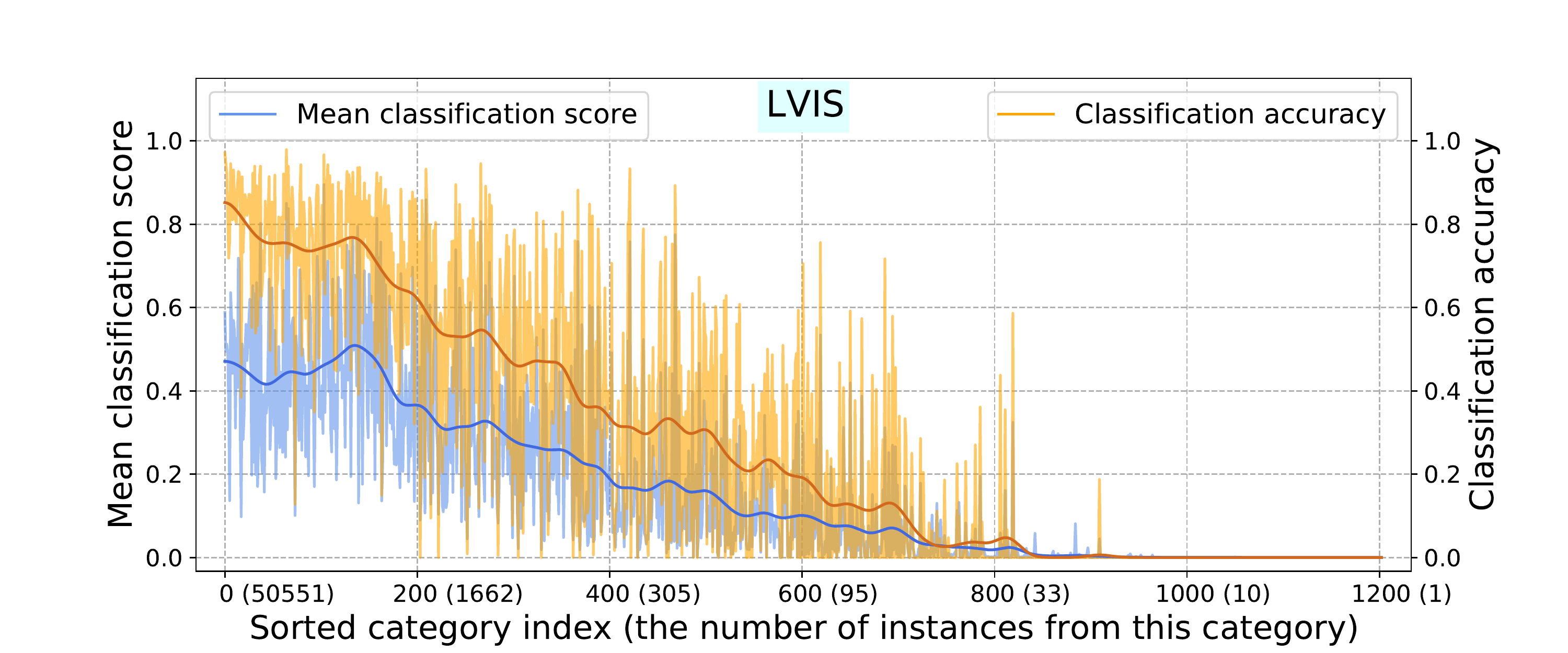}
			\includegraphics[height=3cm,trim=0 0 0 35,clip]{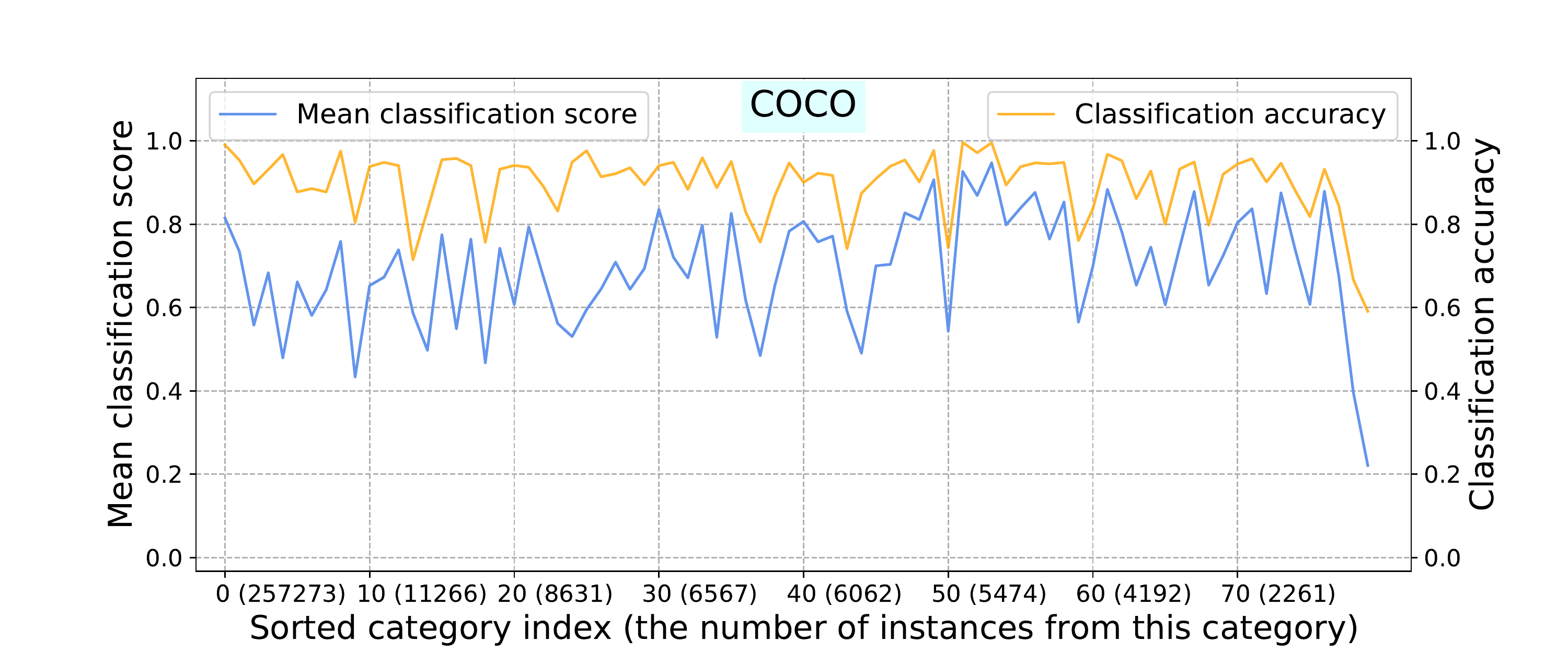}
		\caption{Statistics of mean classification score and classification accuracy for each category on LVIS v1.0 training set and COCO training set tested by Mask R-CNN with ResNet-50-FPN. The x-axis represents the sorted category index and the number of instances from the corresponding category.}
		\label{statistics}
\end{figure}

\newcommand\blfootnote[1]{%
\begingroup 
\renewcommand\thefootnote{}\footnote{#1}%
\addtocounter{footnote}{-1}%
\endgroup 
}
{
	\blfootnote{
	 $^*$Corresponding author.
	 }
}

\section{Introduction} \label{introduction}
Object detection plays an important role in computer vision, and recent object detectors~\cite{he2017mask,lin2017focal,tian2019fcos} have achieved promising performance on several common datasets with a few categories and balanced class distribution, such as PASCAL VOC~(20 classes)~\cite{everingham2010pascal} and COCO~(80 classes)~\cite{lin2014microsoft}. However, most real-world data contains a large number of categories and its distribution is long-tailed: a few head classes contain abundant instances while a great number of tail classes only have a few instances.

Recently, LVIS~\cite{gupta2019lvis} is released for exploring long-tailed object detection. Not surprisingly, the performance of the state-of-the-art detectors designed for balanced data is significantly degraded~\cite{li2020overcoming,wu2020forest} if they are directly applied to such datasets. The reason for the performance degradation mainly comes from two aspects:
(1)~The long-tailed distribution of data. The number of the instances from tail classes~(\eg, only 1 instance for one class) is insufficient for training a deep learning model, resulting in under-fitting of these classes. Moreover, the tail classes will be overwhelmed by the head classes during training, because the number of the instances of head classes is much larger than that of tail classes~(\eg, thousands of times). As a result, the detectors cannot learn the tail classes well, and recognize those tail classes with very low confidence, as demonstrated in Figure~\ref{statistics}.
(2)~The large number of categories. With the increase of the number of categories,
it brings a higher chance of misclassification, especially for the tail classes with a very low classification score.

Several works~\cite{gupta2019lvis,ren2020balanced,tan2020equalization,wang2020devil} attempted to cope with the problem of long-tail learning by re-sampling training data or re-weighting loss function. However, most of them assign the sampling rate and the loss weight according to the sampling frequency of each category, which is model-agnostic and sensitive to hyper-parameters~\cite{li2020overcoming,ren2020balanced}. It may bring the following problems: (1)~the model-agnostic data re-sampling is prone to over-fit the tail classes and under-represent the head classes; (2)~the dataset-based loss re-weighting may cause excessive gradients and unstable training especially when the category distribution is extremely imbalanced. Recently, wang~\etal~\cite{wang2021seesaw} introduce Seesaw loss to adaptively re-balance the gradients of positive and negative samples by dynamically accumulating the number of class instances during training. However, the number of the training samples cannot accurately reflect the learning quality of the classes, due to the diversity and complexity of instances and categories, \eg, training a classifier for the categories with visual similarity usually requires more training samples than the categories with very different visual appearance.

To address the above problems, we propose to use the mean classification score to monitor the learning status~(\ie, classification accuracy) of each category during training. 
As shown in Figure~\ref{statistics}, the mean classification score has an approximate positive correlation with the classification accuracy. Thus, it can be used as an effective indicator to reflect the classification accuracy during training.
Based on this indicator, we design an Equilibrium Loss~(EBL) and a Memory-augmented Feature Sampling~(MFS) method, to dynamically balance the classification.
\textbf{Equilibrium Loss:} To balance the classification of different classes, EBL assigns different loss margins between any two classes based on the statistical mean classification score. It increases the loss margin between weak~(with low mean score) positive classes and dominant~(with high mean score) negative classes, and vice versa. Thus, the designed loss margin increases the intensity of the adjustment of the classification decision boundary for the weak classes, resulting in a more balanced classification.
\textbf{Memory-augmented Feature Sampling:} In addition to increasing the intensity of the adjustment of the decision boundary, we design MFS to increase the frequency and accuracy of the adjustment of the decision boundary for the weak classes. 
Specifically, rich instance features are firstly extracted based on a set of dense bounding boxes generated by a model-agnostic bounding box generator, and then stored by a feature memory module for feature reuse across training iterations. 
Finally, a probabilistic sampler is used to access the feature memory module to sample more instance features of weak classes to improve the training.

We codename the proposed method Long-tailed Object detector with Classification Equilibrium~(LOCE).
In summary, our contributions are as follows: (1)~we propose to use the mean classification score to monitor the classification accuracy of each category during training;
(2)~we develop a score-guided equilibrium loss that improves the intensity of the adjustment of the decision boundary for the weak classes. 
(3)~we design a memory-augmented feature sampling to enhance the frequency and accuracy of the adjustment of the decision boundary for the weak classes.
(4)~we conduct experiments on LVIS~\cite{gupta2019lvis} using Mask R-CNN~\cite{he2017mask} with various backbones including ResNet-50-FPN and ResNet-101-FPN~\cite{he2016deep,lin2017feature}. Extensive experiments show the superiority of LOCE. It improves the tail classes by 15.6 AP based on the Mask R-CNN with ResNet-50-FPN~\cite{he2016deep,lin2017feature} and outperforms the most recent long-tailed object detectors by more than 1 AP on LVIS v1.0.

\begin{figure*}[ht]
\small
\centering
\begin{minipage}{0.28\linewidth}
\centerline{\includegraphics[height=4cm]{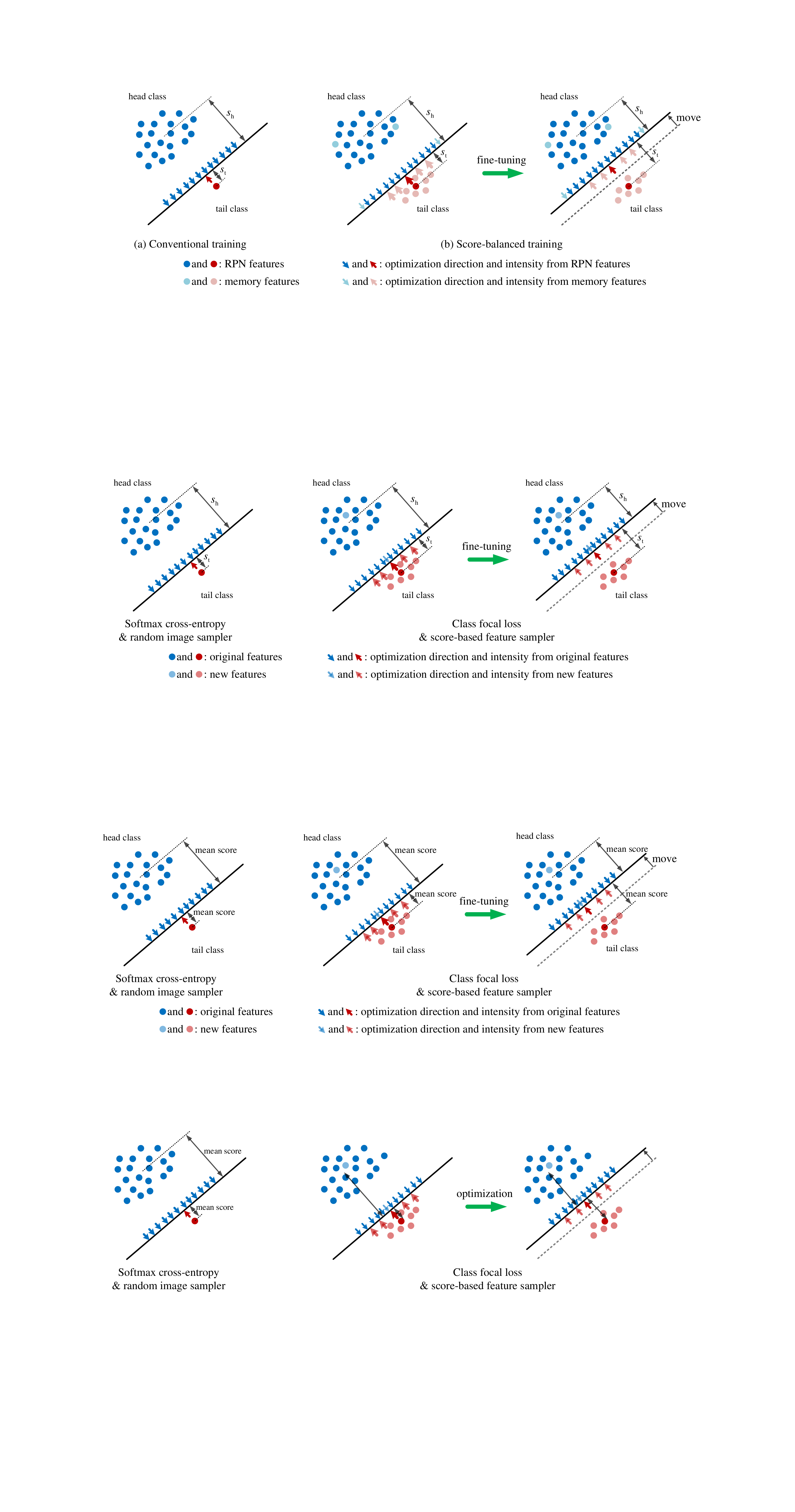}}
\centerline{Conventional training}
\end{minipage}
\qquad \qquad
\begin{minipage}{0.6\linewidth}
\centerline{\includegraphics[height=4cm]{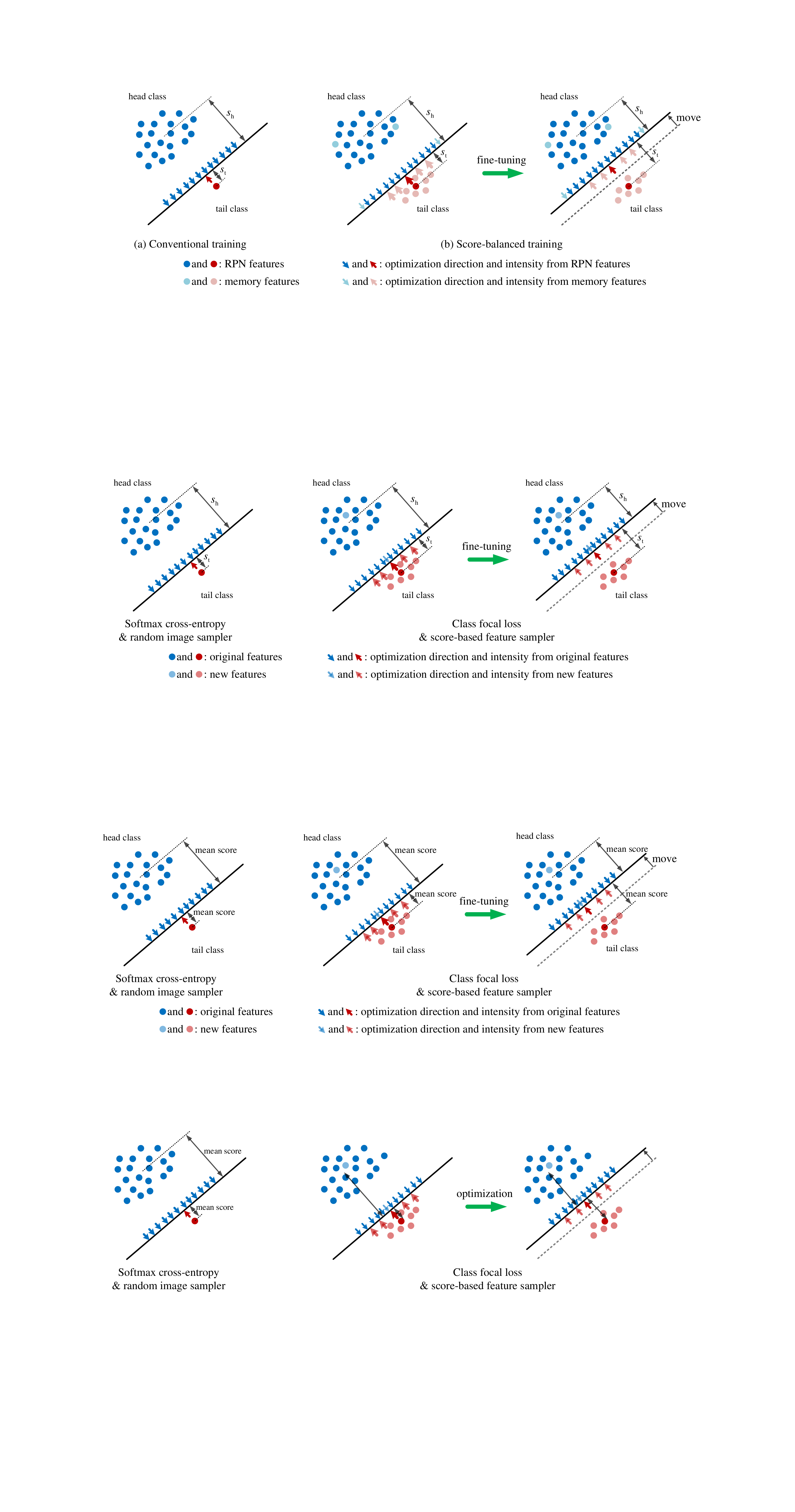}}
\centerline{Score-balanced training}
\end{minipage}
\includegraphics[height=0.76cm]{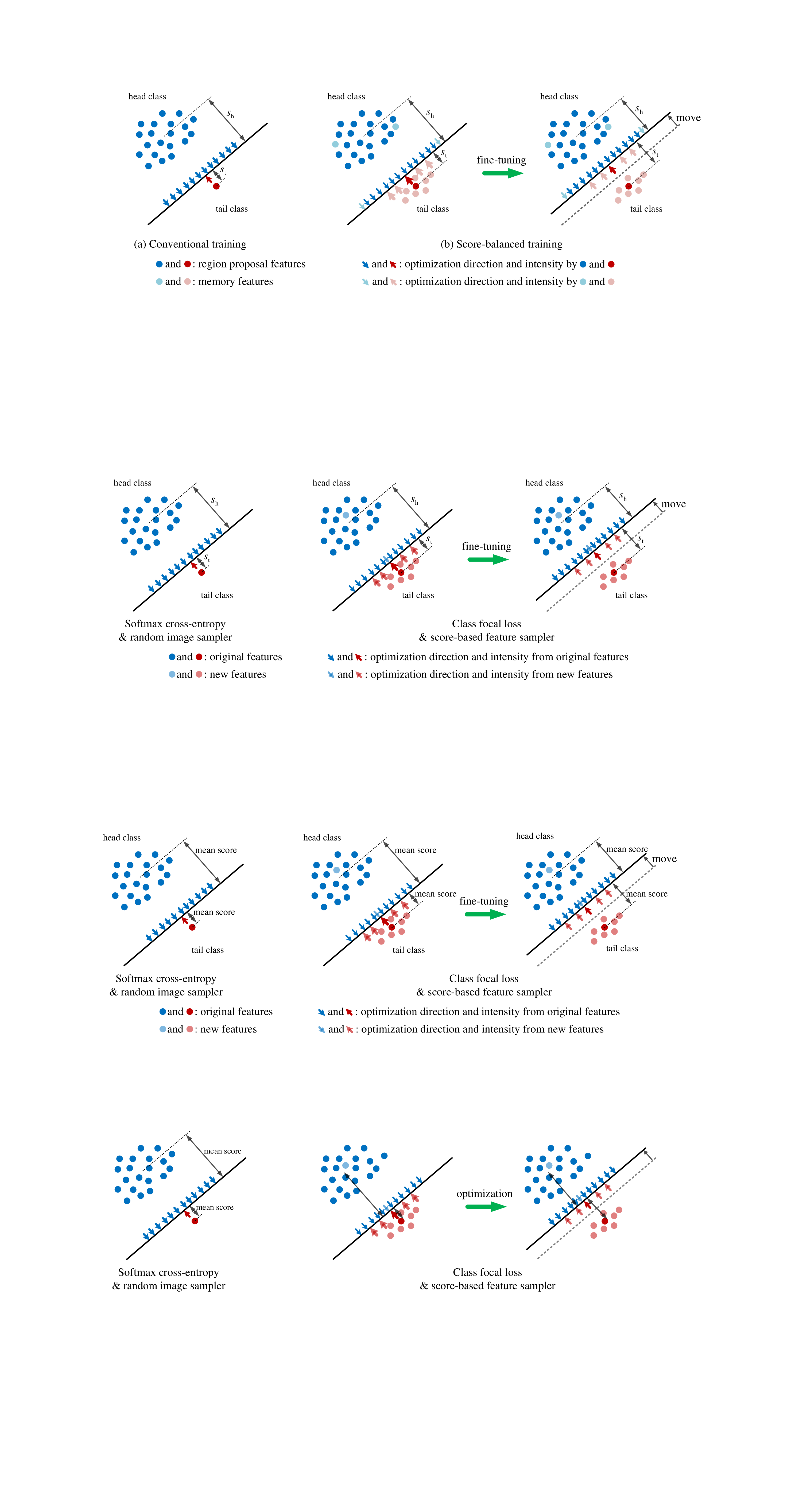}
\caption{Demonstration of the adjustment of the decision boundary for classification balance. 
For simplicity, we only demonstrate the adjustment process between a head class and a tail class. 
The relative magnitudes of the mean classification score can approximately reflect the distances from the class feature center to decision boundary, \ie, $s_{h}$ and $s_{t}$. The region proposal features are obtained from RPN. The memory features are obtained from feature memory which stores rich instance features from the dense bounding boxes. The direction and size of the arrows mean the direction and intensity of the adjustment of the decision boundary. Specifically, EBL~(Section~\ref{section-loss}) decides the size of the arrows while MFS~(Section~\ref{section-sampler}) decides the number of the arrows. With stronger and more adjustments from tail class, the decision boundary moves from tail class to head class until the equilibrium is reached.}
\vspace{-2.5mm}
\label{demo-boundary}
\end{figure*}

\section{Related Work}
\vspace{-1mm}
\paragraph{Object Detection.}
Modern object detection frameworks~\cite{he2017mask,lin2017focal,ren2015faster,tian2019fcos,zhong2020representation} can be divided into two-stage and one-stage ones. The two-stage detectors first generate a set of region proposals, and then classify and refine the proposals. In contrast, the one-stage detectors directly predict the category and bounding box at each location. Most of the detectors are designed for balanced data. When it comes to long-tailed data, the performance of such detectors is greatly degraded. Recently, extensive studies attempted to optimize the two-stage detectors such as~\cite{he2017mask,ren2015faster} to cope with the long-tailed data, by designing balanced samplers~\cite{gupta2019lvis,ren2020balanced,wang2020devil,wu2020forest} or balanced loss functions~\cite{li2020overcoming,ren2020balanced,tan2020equalization,wang2021seesaw}.
Some works~\cite{li2020overcoming,wang2020devil} adopt decoupled training pipeline~\cite{kang2019decoupling}, which first learns the universal representations with unbalanced data and then fine-tunes the classifier with re-balanced data or balanced loss function.
Inspired by them, we propose both adaptive feature sampling and adaptive loss function for long-tailed detection.

\vspace{-3mm}
\paragraph{Sampler for long-tail learning.}
Data re-sampling is a common solution for long-tail learning. It typically over-samples the training data from tail classes while under-samples those from head classes. In long-tailed detection, the data samplers balance the training data on the image-level or instance-level. Gupta~\etal~\cite{gupta2019lvis} use image-level Repeat Factor Sampling~(RFS) to up-sample the data from minority classes based on the sampling frequency of each class. Wang~\etal~\cite{wang2020devil} propose a class-based sampler to balance the data from the instance-level, by only considering the proposals of the selected classes. Wu~\etal~\cite{wu2020forest} set a higher NMS threshold for tail classes to sample more proposals from tail classes. These methods design the balanced sampler depending on the frequency distribution of categories. In contrast, we design a memory-augmented feature sampling based on the mean classification score, which can adapt to the training process dynamically. Recently, ren~\etal~\cite{ren2020balanced} introduce Meta Sampler to estimate the optimal sample rate with meta-learning. Compared with Meta Sampler, our feature sampling is simpler and more versatile.

\paragraph{Loss Function for long-tail learning.}
Balanced loss functions have received lots of attention in long-tailed classification. Most of them are achieved through loss weighting or margin modification related to the distribution of training data. For example, the works such as~\cite{morik1999combining,xie1989logit} re-weight the loss functions by the inverse of the sampling frequency of each class, while those of~\cite{cao2019learning,menon2020long,tan2020equalization} increase the loss margins of tail classes and decrease those of head classes for balanced classification. Recently, several works attempted to design balanced loss for long-tailed detection. Tan~\etal~\cite{tan2020equalization} propose Equalization Loss~(EQL) to improve the performance of tail classes by ignoring the suppressing gradients for tail classes. Li~\etal~\cite{li2020overcoming} introduce Balanced Group Softmax that first groups the classes based on the instance numbers and then separately apply softmax within each group. Ren~\etal~\cite{ren2020balanced} design Balanced Softmax to accommodate the label distribution shifts of the long-tailed data according to the number of category samples. Tan~\etal~\cite{tan2020equalization} improve EQL by re-balancing the positive and negative gradients for each category independently and equally. Wang~\etal~\cite{wang2021seesaw} develop Seesaw loss to re-balance the gradients of positive and negative samples by accumulating the number of the training samples. Different from the existing methods, we design the loss function according to the mean classification score calculated during training. It can track and adjust the learning status of the model dynamically.

\section{Methodology}
\vspace{-1mm}
As mentioned in Section~\ref{introduction}, if the distribution of the training data is severely skewed, the mean classification score obtained by the conventional detectors is extremely imbalanced for each category. In this work, we propose LOCE, an object detector with classification equilibrium, to alleviate this problem. We first use the mean classification score to indicate the learning status~(\ie, classification accuracy) of each category during training~(Section~\ref{section-preliminary}). Then, based on this indicator, we balance the classification through a score-guided equilibrium loss~(Section~\ref{section-loss}) and a memory-augmented feature sampling method~(Section~\ref{section-sampler}).
The proposed loss function and the feature sampling method collaboratively adjust the classification decision boundary, as demonstrated in Figure~\ref{demo-boundary}.
Similar to most methods~\cite{kang2019decoupling,li2020overcoming,wang2020devil} for long-tailed object classification or detection, we adopt the decoupled training pipeline~\cite{kang2019decoupling}. The two methods are adopted in the fine-tuning stage.

\subsection{Mean Classification Score} \label{section-preliminary}
We first analyze the classification problem when applying conventional detectors to long-tailed data, and then introduce the mean classification score to indicate the learning status of the detector.
\vspace{-3mm}
\paragraph{Problem formulation.} Most conventional detectors designed for balanced data are optimized with softmax cross-entropy loss function and random image sampler. Formally, let $F$ denotes a scorer that takes an image $x$ as input and generates a category prediction $z = F(x)$, where $z \in \mathbb{R}^{C+1}$ ($C$ object classes plus a background class). During training, with the corresponding ground truth label $y \in \{1, 2, ..., C + 1\}$, the standard softmax cross-entropy can be written as:
\begin{equation}
	\resizebox{.90\linewidth}{!}{$
    \displaystyle
	L(z, y) = -log \frac{e^{z_{y}}}{\sum_{y' \in \{1, 2, ..., C + 1\}}e^{z_{y'}}}
	= log[1 + \sum_{y' \neq y} e^{z_{y'} - z_{y}}].
	$}
\end{equation}
With a random image sampler, each image $x$ is selected with equal probability during training. Considering the long-tail learning where the distribution $\mathbb{P}(y)$ is highly skewed, the images from tail classes have a very low probability of occurrence. Under the optimization of softmax cross-entropy and random image sampler, the classifier tends to predict unbalanced classification scores resulting in lots of misclassification for tail classes, as demonstrated in Figure~\ref{statistics}.

\vspace{-3mm}
\paragraph{Classification accuracy indicator.} To alleviate the problem of classification imbalance, we attempt to find an effective indicator to reflect the learning status~(\ie, classification accuracy) of the classifier for each category and dynamically adjust the learning process.
Previous works~\cite{cao2019learning,gupta2019lvis,menon2020long,tan2020equalization,wu2020forest} proposed to balance the classification based on the number of training samples of each category. However, the number of the training samples cannot indicate the learning quality of the model accurately, because of the diversity and complexity of instances and categories. For example, training a classifier for the categories with high inter-class visual similarity usually requires more training samples than the categories with very different visual appearances.
Instead, we seek a more effective indicator to reflect the classification status.
From the statistics shown in Figure~\ref{statistics}, we found that the mean classification score has an approximate positive correlation with the classification performance.
Namely, for LVIS, the head classes have higher mean classification scores and higher classification accuracy, while the tail classes have lower mean classification scores and lower classification accuracy. For the balanced dataset COCO, we also observe a similar pattern: high mean classification scores are usually associated with high classification accuracy.

Therefore, we propose to use the mean classification score to indicate the learning status of the model for each category.
However, computing the mean classification score for the whole dataset at each training iteration is infeasible.
Instead, we approximate the mean classification score by a mean score vector $s \in \mathbb{R}^{C+1}$ during training.
For an instance with positive label $y$ at the $i$th iteration, we update the corresponding $s_{y}^{i}$ in the mean score vector with the exponential moving average function:
	\begin{equation}
	s_{y}^{i} = \alpha s_{y}^{i - 1} + (1 - \alpha) p_{y}^{i},
	\label{eq-score}
	\end{equation}
where $p_{y}^{i}$ is the predicted probability of the instance, and $\alpha$ is a smoothing coefficient hyper-parameter. 

Compared with the existing works~\cite{ren2020balanced,wang2021seesaw} that use the number of training instances to indicate the learning status of the model for each category, the proposed indicator has the following advantages: (1)~it can monitor the classification accuracy of each category during training;
(2)~it can be applied when the distribution of the training data is not visible or the model is pre-trained with other datasets, \eg, the training samples are obtained from an online stream. The mean classification score affects the training by guiding the proposed loss function and the feature sampling method to balance the classifier, which is introduced in the following.

\begin{figure*}[ht]
		\centering
		\includegraphics[height=5.5cm]{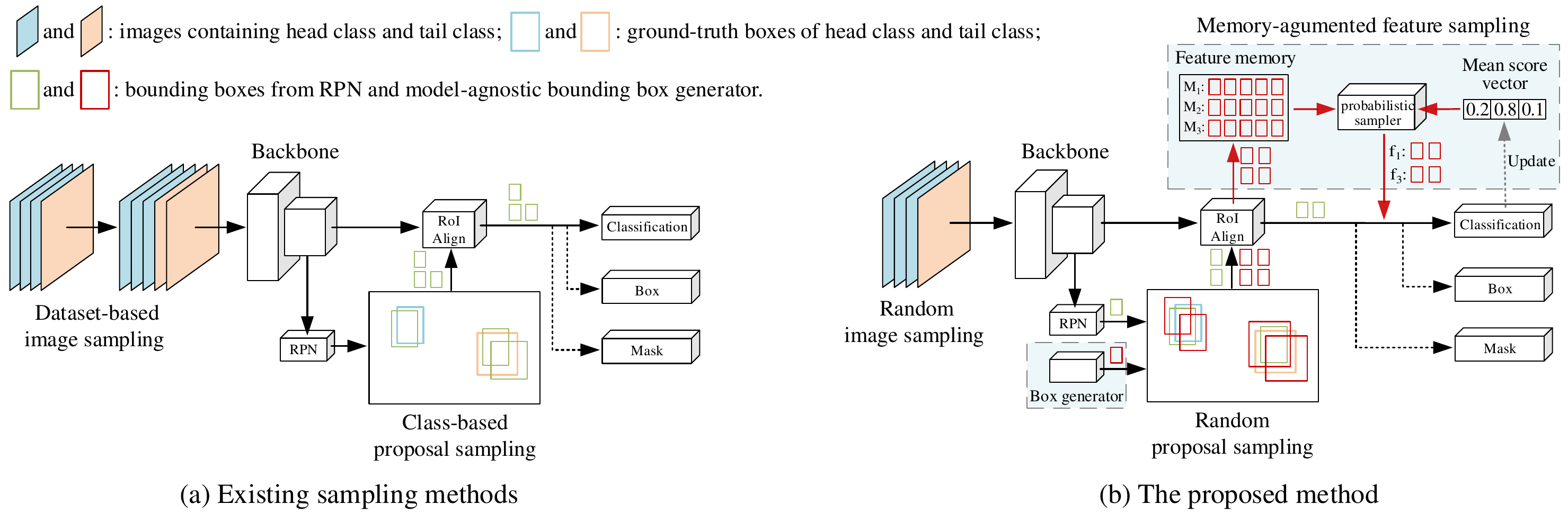}
		\caption{Comparison between the existing balanced sampling methods and the proposed memory-augmented feature sampling method. (a): Dataset-based image sampling~(\eg, RFS~\cite{gupta2019lvis} and CBS~\cite{wang2020devil}) over-samplings the images from tail classes or under-samplings the images from head classes.
		Class-based proposal sampling~(\eg, NMS Resampling~\cite{wu2020forest} and Bi-level Sampling~\cite{wang2020devil}) samples more proposals from tail classes or the selected classes. (b): Memory-augmented feature sampling stores the instance features from the model-agnostic dense bounding boxes by the feature memory module, and samples the memory features based on the mean classification score.}
		\label{demon-sampling}
		\vspace{-3mm}
\end{figure*}

\subsection{Equilibrium Loss} \label{section-loss}
In this section, we introduce the Equilibrium Loss~(EBL) to balance the classification through shifting the decision boundary of the classifier. Recently, several works such as~\cite{cao2019learning,tan2020equalization,wang2021seesaw} also attempt to adjust the decision boundary to balance the classification, based on the prior distribution $\mathbb{P}(y)$ or the accumulated number of the training samples of each category. 
However, as mentioned in Section~\ref{section-preliminary}, the number of the training samples cannot indicate the learning quality of the model accurately.
Different from the existing methods, EBL adjusts the decision boundary according to the mean classification score.
Concretely, if a category~(\eg, a tail class) has a low classification score, we expect EBL to push the decision boundary away from that category and towards other categories~(\eg, head classes). This behavior, in turn, improves the mean classification score of the current category resulting in a more balanced classification.
To achieve that, we add a score-relevant margin $\delta_{yy'}$ into the softmax cross-entropy loss, turning into EBL as:
 	\begin{equation}
	L(z, y) = log[1 + \sum_{y' \neq y} e^{z_{y'} - z_{y} + \delta_{yy'}}],
	\label{eq-loss}
	\end{equation}
where $\delta_{yy'}$ works as a tunable balancing margin between any two classes, based on the distribution of the mean classification score. To adjust the decision boundary accordingly, the design of $\delta_{yy'}$ should satisfy the following two properties:
(1)~it should reduce the suppression of dominant classes~(\ie, having high mean classification score) over weak classes~(\ie, having low mean classification score), which can be achieved by reducing the margin between dominant positive classes and weak negative classes;
(2)~it should enlarge the suppression of weak classes over dominant classes, which can be achieved by increasing the margin between weak positive classes and dominant negative classes. Therefore, we design the following adaptive loss margin between any two classes:
	\begin{equation}
	\delta_{yy'} = log(\frac{s_{y'}}{s_{y}}).
    \label{eq-margin}
	\end{equation}
\vspace{-7mm}
\paragraph{Background class.} As defined above, the background is regarded as an auxiliary category in the classifier. In the experiments, we found that the classifier trained with EBL tends to predict false positive results, \ie, misclassifying the backgrounds as foregrounds. To reduce those false positive cases, we enlarge the corresponding punishment. In the margin loss, it can be achieved by increasing the margin between the positive background class and the negative foreground classes.
Consequently, we decrease the mean classification score of background class $s_{C + 1}$ when computing Eq.(\ref{eq-margin}). It is worth noting that most training samples~(at least 75\%) for classifier are negative samples~(\ie, background). For simplicity and efficiency, instead of computing the statistical $s_{C + 1}$, we use a small value~(\eg, 0.01) to replace $s_{C + 1}$, namely $\hat{s}_{C + 1}$. Some works such as~\cite{wang2021seesaw} attempt to reduce the false positive cases by introducing an extra objectness branch. Comparatively, the proposed method is simpler and more efficient.

\subsection{Memory-augmented Feature Sampling} \label{section-sampler}
\vspace{-1mm}
Although EBL tends to move the decision boundary from the tail classes to the dominant ones, the decision boundary sometimes is still closer to the tail classes.
This is because EBL only adjusts the intensity of the adjustment of the decision boundary, while the frequency of the adjustment is very low for tail classes during training, due to the overwhelming number of images in the head classes.
Especially, the tail classes usually have very few training samples~(\eg, $< 10$ for LVIS). 
Therefore, we need to increase the number (or occurrence) of the training samples of tail classes to enhance the frequency of such boundary adjustment, in addition to EBL.

To increase the occurrence of the training samples from tail classes, a straightforward method is data re-sampling. As shown in Figure~\ref{demon-sampling}(a), the widely used sampling methods for data balancing can be divided into two categories: (1)~Dataset-based image sampling such as Repeat Factor Sampling~(RFS)~\cite{gupta2019lvis} and Class-balanced Sampling~(CBS)~\cite{wang2020devil} usually samples more training images of the tail classes based on the training set statistics. (2)~Class-based proposal sampling such as NMS Resampling\cite{wu2020forest} and Bi-level Sampling~\cite{wang2020devil} samples more proposals from Region Proposal Network~(RPN) for the tail classes or the selected classes. Despite their success, these sampling methods have the following limitations:
(1)~the diversity of the training features from proposal sampling depends on the behavior of RPN;
(2)~image sampling usually requires more training iterations;
(3)~most existing sampling methods are model-agnostic and prone to over-fit the tail classes and under-represent the head classes. To overcome these limitations, we propose a more efficient Memory-augmented Feature Sampling~(MFS) method. As shown in Figure~\ref{demon-sampling}(b), MFS consists of a bounding box generator, a feature memory module, and a probabilistic sampler.

\vspace{-3mm}
\paragraph{Bounding box generator.} Recent two-stage detectors sample RoI features for the classifier based on the proposals from RPN and the sampling configuration. In such pipeline, the diversity of the classification training features is limited by the performance of RPN and the sampling configuration such as the number of positive samples at each iteration.
It hinders the classifier from improving its generalization ability and accuracy, especially for the tail classes.
Instead, we design a model-agnostic bounding box generator with aim to extract rich instance features for training the classifier. 
Moreover, we propose to reuse the extracted instance features across training iterations.
With such design, the feature diversity for the classifier is no longer sensitive to the performance of RPN and the sampling configuration.

Concretely, given an object instance, we have its ground-truth class $y$ and ground-truth box $b = [x_{1}, y_{1}, x_{2}, y_{2}]$, where $(x_{1}, y_{1})$ and $(x_{2}, y_{2})$ are the coordinates of the upper left corner and the lower right corner of the ground-truth box. The bounding box generator yields the following dense bounding boxes:
\vspace{-0.5mm}
	\begin{equation}
	\hat{b} = [x_{1} \pm \frac{\eta_{1} w}{6}, y_{1} \pm \frac{\eta_{2} h}{6}, x_{2} \pm \frac{\eta_{3} w}{6}, y_{2} \pm \frac{\eta_{4} h}{6}],
	\end{equation}
where $w$, $h$ are box width and height, \ie, $w = x_{2} - x_{1}$, $h = y_{2} - y_{1}$, and $\eta_{i} \in [0, 1]$ is a random number. Such bounding box generator can obtain any potential positive bounding boxes~(\ie, IoU $>$ 0.5). Based on the dense bounding boxes $\hat{b}$, we extract the corresponding instance features $f_{y}$ by applying RoI-Align~\cite{he2017mask} to the features from Feature Pyramid Network~(FPN).

\paragraph{Feature Memory Module.}
Image re-sampling usually requires extra training iterations while proposal sampling does not provide enough samples to balance the classification, especially for tail classes.
In contrast, the proposed bounding box generator can yield dense bounding boxes for extracting more instance features, without the requirement of extra training iterations.
However, using all the instance features from the dense bounding boxes to train the classifier within an iteration can bring lots of computational overhead and memory consumption, especially when there are lots of instances in an image.
Besides, for the tail classes that only occur a few times in a training epoch, taking all the instance features within an iteration may still be inadequate.

To alleviate the above problems, we use a feature memory module to store the instance features, and reuse the instance features as needed during the following training.
The rationale is that the parameters of the backbone~(\eg, ResNet-50-FPN) are frozen during the fine-tuning stage.
Thus, the features extracted from the backbone are stable and can be reused to fine-tune the classifier during different training iterations.
Memory module is widely adopted in contrastive learning~\cite{chen2020improved,he2020momentum,li2019memory,wang2020cross}, but never used in long-tailed object detection. 
Specifically, the feature memory module is maintained and updated with a class queue for each class $y$:
	\begin{equation}
	\mathbb{M}_{y} = [f_{y}^{1}, f_{y}^{2}, ... , f_{y}^{M}],
	\end{equation}
where $f_{y}^{i}$ is the $i$th RoI features of class $y$ in the memory module $\mathbb{M}_{y}$, and $M$ is the memory size. At each iteration, we enqueue the instance features $f_{y}$ of current mini-batch into the corresponding class queue, and dequeue the instance features of the earliest mini-batch accordingly. Most two-stage detectors such as Mask R-CNN share Fully-connected~(FC) layers for classification and box regression before the prediction layers. We also store the regression target $t_{y}^{i}$ corresponding to $f_{y}^{i}$ to train the box branch, which can improve the accuracy of box regression for tail classes with negligible overhead.
	
\paragraph{Probabilistic Sampler.} At each training iteration, we access the feature memory module by a sampler to augment the training features for balancing the classifier. 
In particular, we use the mean classification score to adaptively adjust the sampling process, similar to that for EBL. To improve the training effectiveness and classification score of the weak classes~(\eg, tail classes), we sample more memory features of these classes. Specifically, we design a probabilistic sampler that samples the memory features according to the probability $p$ with negative correlation with the mean classification score, namely:
	\begin{equation}
	p_{y} = \frac{f(s_{y})}{\sum_{y'}f(s_{y'})},
	\end{equation}
	where $f(\cdot)$ is a non-increasing transform and $p_{y}$ is the sampling probability of class $y$. For simplicity, we define:
	\vspace{-0.5mm}
	\begin{equation}
	f(s_{y}) = \frac{1}{s_{y}}.
	\vspace{-0.5mm}
	\end{equation}
Finally, we randomly choose $k$ classes according to $p$, and select $m$ features from the feature memory for each selected class. Then, the $k \times m$ selected features will be used together with the RoI features from RPN to train the classifier.

\section{Experiments}
\subsection{Experimental Setup}
\vspace{-1mm}
\paragraph{Dataset and evaluation metric.} We conduct experiments on the recent long-tail and large-scale dataset LVIS~\cite{gupta2019lvis}. The latest version v1.0 contains 1203 categories with both bounding box and instance mask annotations. We use the $train$ set~(100k images with 1.3M instances) for training and $val$ set~(19.8k images) for validation.
We also perform experiments on LVIS v0.5, which contains 1230 and 830 categories respectively in its $train$ set and $val$ set.
All the categories are divided into three groups based on the number of the images that each category appears in the $train$ set: rare~(1-10 images), common~(11-100 images), and frequent~($>$100 images). Apart from the official metrics Average Precision~(AP), we also report AP$_{r}$~(for rare classes), AP$_{c}$~(for common classes) and AP$_{f}$~(for frequent classes) to measure the detection performance and segmentation performance. 
Unless specific, AP$^{b}$ denotes the detection performance, while AP denotes the segmentation performance.

\vspace{-3mm}
\paragraph{Implementation details.}
We implement our method with MMDetection~\cite{chen2019mmdetection} and conduct experiments using Mask R-CNN~\cite{he2017mask} with various backbones including ResNet-50-FPN and ResNet-101-FPN~\cite{he2016deep,lin2017feature} pre-trained on ImageNet~\cite{deng2009imagenet}. Following~\cite{ren2020balanced}, we use the decoupled training pipeline. Namely, we first train the model with standard softmax cross-entropy and random image sampler for 24 epochs, then fine-tune the model with the proposed method for 6 epochs. Specifically, the initial learning rate is  0.02 and dropped by a factor of 10 at the $16$th and $22$th epoch for the first training stage and the $3$th and $5$th epoch for the fine-tuning stage. The models are trained using SGD optimizer with 0.9 momentum and 0.0001 weight decay and batch size of 16 on 8 GPUs. Following the convention, we train the detector with scale jitter~(640-800) and horizontal flipping. At testing time, the model is evaluated without test time augmentation, and the maximum number of detections per image is 300 with the minimum score threshold of 0.0001. We set $\alpha$ as 0.9 for updating the mean classification score. The substituted $\hat{s}_{C + 1}$ is set to be 0.01. The memory size $M$ is  80, and $k$ and $m$ in feature sampler are  8 and 4 on each GPU.
As in~\cite{wang2021seesaw}, we adopt normalized linear activation to the mask prediction. More implementation and training details refer to the supplementary material.

\begin{table}
\small
\centering
\begin{tabular}{cp{0.5cm}>{\centering}p{0.55cm}>{\centering}ccccc}
\toprule
Indicator & EBL & MFS & AP$^{b}$ & AP & AP$_{r}$ & AP$_{c}$ & AP$_{f}$ \\
\cmidrule(r){1-1}
\cmidrule(r){2-3}
\cmidrule(r){4-5}
\cmidrule(r){6-8}
$s$ & & & 20.4 & 20.2 & 2.9 & 18.2 & 30.1 \\
& $\checkmark$ & & 24.0 & 23.8 & 8.3 & 23.7 & 30.7 \\
&  & $\checkmark$ & 26.0 & 25.7 & 18.4 & 25.0 & 29.8 \\
& $\checkmark$ & $\checkmark$ & \textbf{27.4} & \textbf{26.6} & 18.5 & \textbf{26.2} & \textbf{30.7} \\
\cmidrule(r){1-1}
\cmidrule(r){2-3}
\cmidrule(r){4-5}
\cmidrule(r){6-8}
$\mathbb{P}(y)$ & $\checkmark$ & $\checkmark$ & 23.4 & 23.1 & \textbf{20.8} & 23.1 & 24.1 \\
\bottomrule
\end{tabular}
\caption{Effectiveness of each proposed component. $s$ denotes the mean classification score, and $\mathbb{P}(y)$ denotes the class distribution of the dataset.}
\label{table-component}
\vspace{-1mm}
\end{table}

\begin{table}
\small
\centering
\begin{tabular}{cccccc}
\toprule
$\alpha$ & AP$^{b}$ & AP & AP$_{r}$ & AP$_{c}$ & AP$_{f}$ \\
\cmidrule(r){1-1}
\cmidrule(r){2-3}
\cmidrule(r){3-6}
0.8 & 27.2 & 26.4 & 17.2 & 26.3 & 30.7 \\
0.9 & \textbf{27.4} & \textbf{26.6} & \textbf{18.5} & 26.2 & \textbf{30.7} \\
0.95 & 27.3 & 26.6 & 18.1 & \textbf{26.4} & 30.7 \\
\bottomrule
\end{tabular}
\caption{Analysis of different smoothing coefficients of $\alpha$.}
\label{table-smoothing}
\vspace{-1mm}
\end{table}

\begin{table}
\small
\centering
\begin{tabular}{cccccc}
\toprule
$\hat{s}_{C+1}$ & AP$^{b}$ & AP & AP$_{r}$ & AP$_{c}$ & AP$_{f}$ \\
\cmidrule(r){1-1}
\cmidrule(r){2-3}
\cmidrule(r){3-6}
0.1 & 26.7 & 26.0 & 17.8 & 25.5 & 30.3 \\
0.01 & \textbf{27.4} & \textbf{26.6} & \textbf{18.5} & \textbf{26.2} & \textbf{30.7} \\
0.001 & 27.4 & 26.5 & 18.4 & 26.0 & 30.7 \\
\bottomrule
\end{tabular}
\caption{Analysis of different value of $\hat{s}_{C+1}$ in equilibrium loss.}
\label{table-bgscore}
\vspace{-1mm}
\end{table}

\begin{table}
\small
\centering
\begin{tabular}{ccccccc}
\toprule
$k$ & $m$ & AP$^{b}$ & AP & AP$_{r}$ & AP$_{c}$ & AP$_{f}$ \\
\cmidrule(r){1-2}
\cmidrule(r){3-4}
\cmidrule(r){5-7}
2 & 2 & 27.1 & 26.3 & 17.8 & 26.1 & 30.6\\
4 & 4 & 27.2 & 26.5 & 18.3 & 26.1 & 30.6 \\
4 & 8 & 27.3 & 26.6 & 18.5 & \textbf{26.3} & 30.6 \\
8 & 4 & \textbf{27.4} & \textbf{26.6} & \textbf{18.5} & 26.2 & \textbf{30.7} \\
8 & 8 & 27.2 & 26.5 & 17.5 & 26.3 & 30.7 \\
\bottomrule
\end{tabular}
\caption{Analysis of different settings of $k$ and $m$ in sampler.}
\label{table-sampler}
\vspace{-3mm}
\end{table}

\begin{figure}[ht]
		\centering
		\subfigure[Softmax cross-entropy loss + random image sampling.]{
			\includegraphics[height=3cm,trim=0 0 0 35,clip]{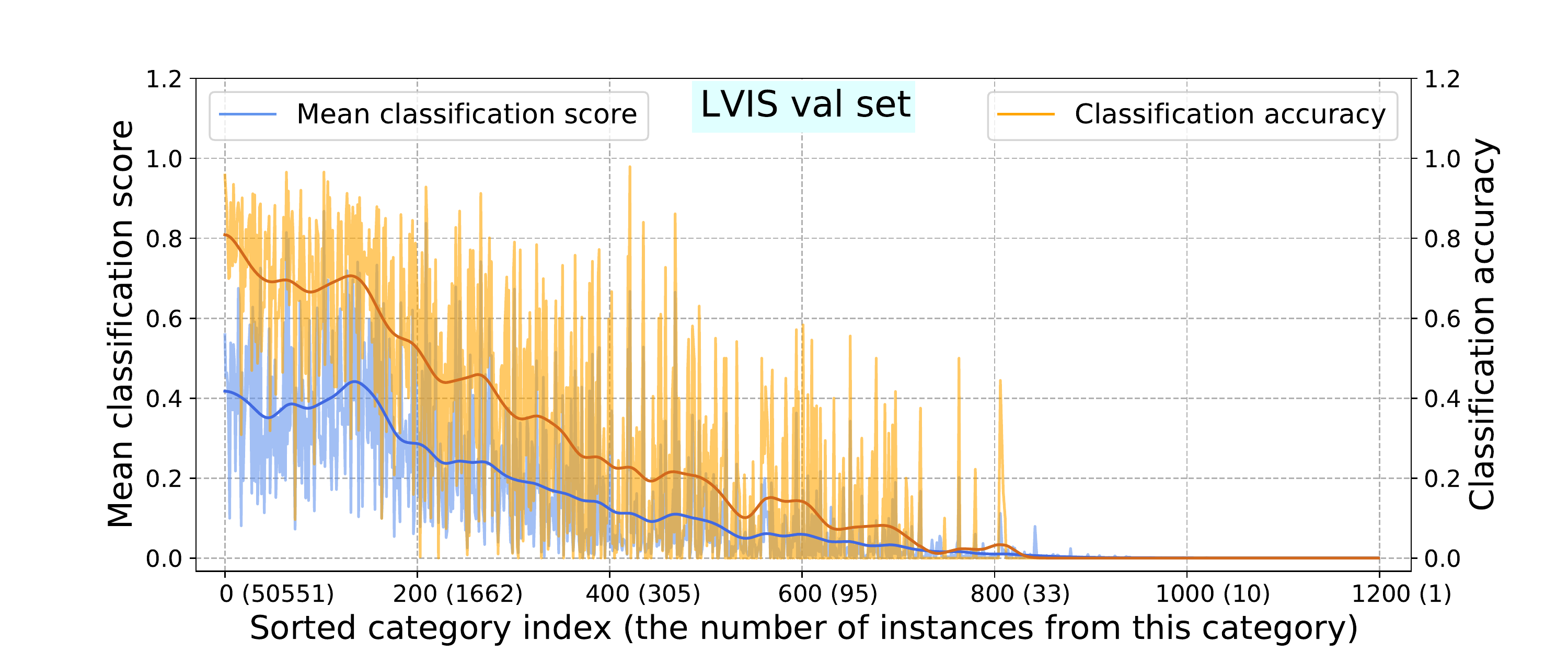}
		}
		\subfigure[Softmax cross-entropy loss + RFS.]{
			\includegraphics[height=3cm,trim=0 0 0 35,clip]{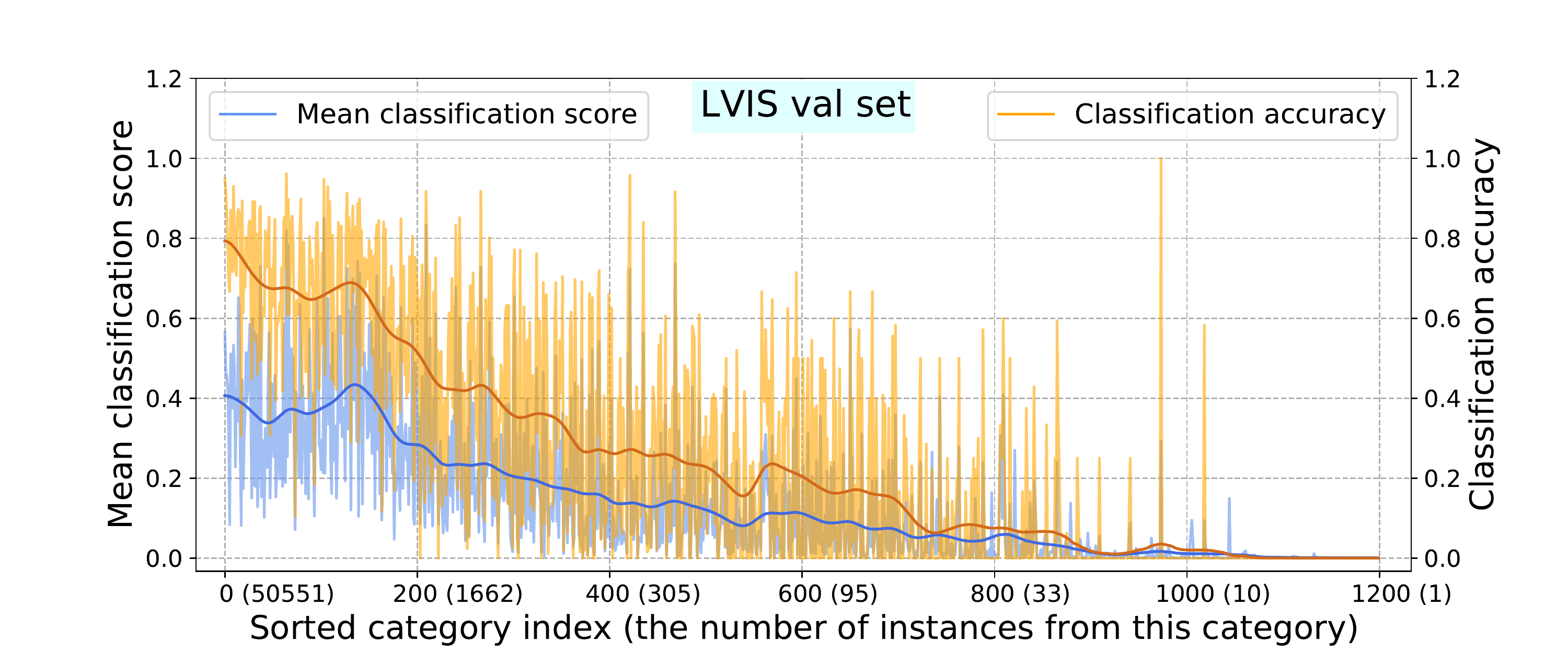}
		}
		\subfigure[LOCE (EBL + MFS).]{
			\includegraphics[height=3cm,trim=0 0 0 35,clip]{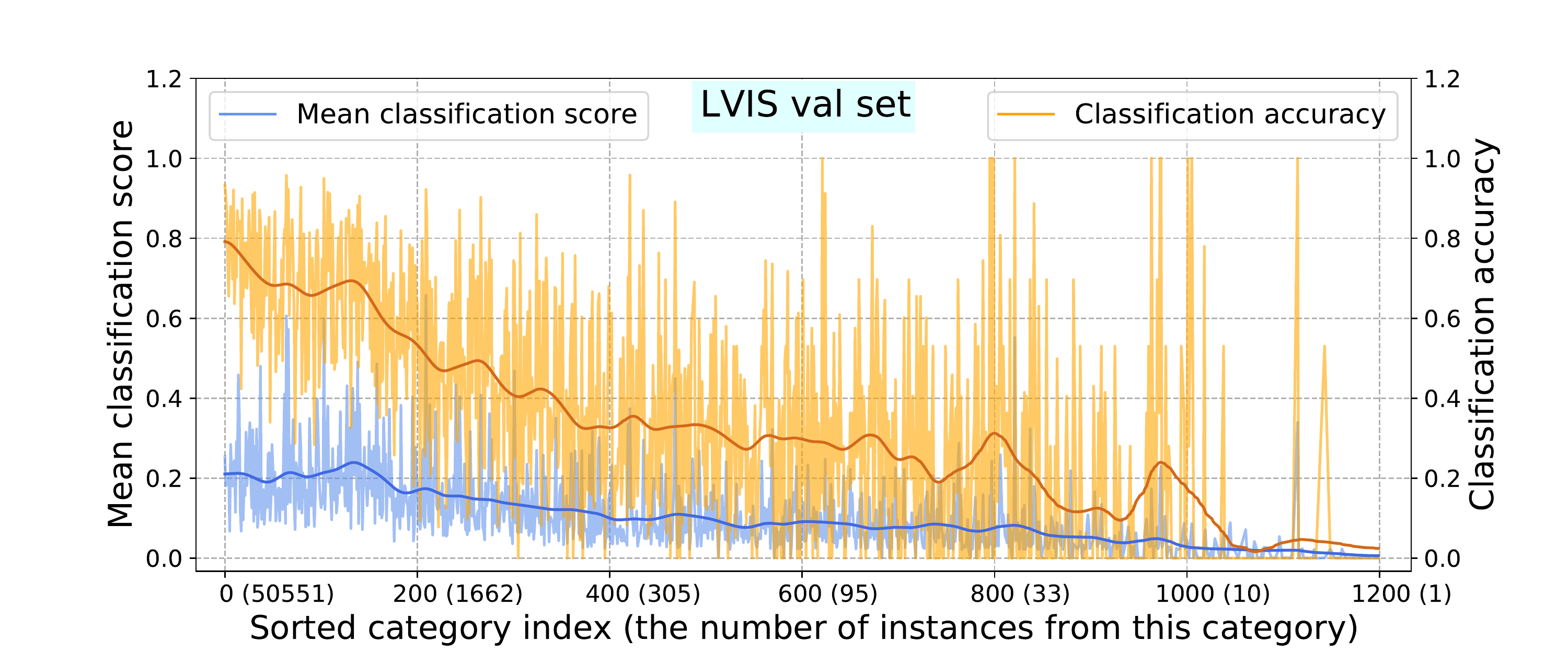}
		}
		\caption{Analysis of classification equilibrium between different training methods on LVIS v1.0 val set. 
		The x-axis represents the sorted category index obtained from the training set.}
		\label{table-analysis}
		\vspace{-3mm}
\end{figure}

\subsection{Ablation Study}
We use Mask R-CNN with backbone ResNet-50-FPN for ablation study, and report the results on LVIS v1.0.

\vspace{-3mm}
\paragraph{Component analysis.} Table~\ref{table-component} reports the detection and segmentation results of each proposed component. For a fair comparison, we train the baseline using standard softmax cross-entropy and random image sampler for 30 epochs. First, we evaluate the performance of EBL. EBL improves both the AP$^{b}$ and AP by 3.6 AP, comparing to the baseline. Specifically, it improves the performance of the classes of all groups, \ie, +5.4 AP for rare classes, +5.5 AP for common classes and +0.6 AP for frequent classes, respectively. These results show that the score-guided loss margin can also help to optimize the decision boundary of head classes, even though it is mainly designed for weak classes. We then examine the effectiveness of MFS. Compared with the baseline, MFS improves the performance by 5.6 AP for object detection and 5.3 AP for instance segmentation. To be more specific, most of the improvements are from rare classes and common classes, which yield +15.5 AP and +6.8 AP improvement for instance segmentation. We can see that using more instance features from weak classes can bring large improvements, especially for rare classes. Next, we verify the effectiveness of the complete method~(\ie, LOCE). EBL and MFS work collaboratively and improve the AP by 7.0 AP for object detection and 6.4 AP for instance segmentation, comparing to the baseline. Notably, MFS itself achieves a little lower performance on frequent classes than the baseline while LOCE achieves higher performance on frequent classes.
This shows that EBL helps MFS to find a better equilibrium point for the frequent classes. Therefore, LOCE dramatically improves the performance of tail classes, while maintaining or even improving the performance of head classes.

\begin{table*}
\small
\centering
\begin{tabular}{lcccccccc}
\toprule
Method & Framework & Backbone & Dataset & AP$^{b}$ & AP & AP$_{r}$ & AP$_{c}$ & AP$_{f}$ \\
\cmidrule(r){1-1}
\cmidrule(r){2-3}
\cmidrule(r){4-4}
\cmidrule(r){5-6}
\cmidrule(r){7-9}
RFS~\cite{gupta2019lvis} & \multirow{6}{*}{Mask R-CNN} & \multirow{6}{*}{R-50-FPN} & \multirow{6}{*}{LVIS v0.5} & 26.1 & 25.9 & 17.8 & 26.2 & 28.8 \\
EQL~\cite{tan2020equalization} &  &  &  & 24.1 & 25.2 & 14.6 & 24.4 & 26.8 \\
Forest R-CNN~\cite{wu2020forest} &  &  &  & 25.9 &  25.6 & 18.3 & 26.4 & 27.6 \\
BAGS~\cite{li2020overcoming} &  &  &  & 25.8 & 26.3 & 18.0 & 26.9 & 28.7 \\
BALMS~\cite{ren2020balanced} &  &  &  & 27.6 & 27.0 & 19.6 & 28.9 & 27.5 \\
EQL v2~\cite{tan2021equalizationv2}$^{\dag}$ &  &  &  & 27.0 & 27.1 & 18.6 & 27.6 & 29.9 \\
\textbf{LOCE} (ours) &  &  &  & \textbf{28.2}& \textbf{28.4} & \textbf{22.0} & \textbf{29.0} & \textbf{30.2} \\
\cmidrule(r){1-1}
\cmidrule(r){2-3}
\cmidrule(r){4-4}
\cmidrule(r){5-6}
\cmidrule(r){7-9}
RFS~\cite{gupta2019lvis} & \multirow{5}{*}{Mask R-CNN} & \multirow{5}{*}{R-50-FPN} & \multirow{5}{*}{LVIS v1.0} & 24.7 & 23.7 & 13.5 & 22.8 & 29.3 
\\
EQL~\cite{tan2020equalization} &  &  &  & 22.5 & 21.6 & 3.8 & 21.7 & 29.2 \\
Seesaw Loss~\cite{wang2021seesaw}$^{*}$ $^{\dag}$ &  &  &  & 24.3 & 23.3 & 13.0 & 22.9 & 28.2 \\
EQL v2~\cite{tan2021equalizationv2}$^{\dag}$ &  &  &  & 26.1 & 25.5 & 17.7 & 24.3 & 30.2 \\
\textbf{LOCE} (ours) &  &  &  & \textbf{27.4} & \textbf{26.6} & \textbf{18.5} & \textbf{26.2} & \textbf{30.7} \\
\cmidrule(r){1-1}
\cmidrule(r){2-3}
\cmidrule(r){4-4}
\cmidrule(r){5-6}
\cmidrule(r){7-9}
RFS~\cite{gupta2019lvis} & \multirow{6}{*}{Mask R-CNN} & \multirow{6}{*}{R-101-FPN} & \multirow{6}{*}{LVIS v1.0} & 26.6 & 25.5 & 16.6 & 24.5 & 30.6 \\
EQL~\cite{tan2020equalization} &  &  &  & 24.0 & 22.7 & 3.7 & 23.3 & 30.4 \\
BAGS~\cite{li2020overcoming} &  &  &  & 26.4 & 25.6 & 17.3 & 25.0 & 30.1 \\
Seesaw Loss~\cite{wang2021seesaw}$^{\dag}$ &  &  &  & 27.4 & 27.1 & 18.7 & 26.3 & 31.7 \\
EQL v2~\cite{tan2021equalizationv2}$^{\dag}$ &  &  &  & 27.9 & 27.2 & \textbf{20.6} & 25.9 & 31.4 \\
\textbf{LOCE} (ours) &  &  & & \textbf{29.0} & \textbf{28.0} & 19.5 & \textbf{27.8} & \textbf{32.0} \\
\bottomrule
\end{tabular}
\caption{Comparison with the state-of-the-art on LVIS v0.5 and v1.0. $^{*}$ indicates 1x training schedule. $^{\dag}$ indicates the concurrent work.}
\label{table-comparison}
\vspace{-2mm}
\end{table*}

Lastly, we evaluate the performance of using the prior distribution of the dataset~(\ie, $\mathbb{P}(y)$) to guide EBL and MFS to balance the classification.
Specifically, we use the number of the instances of each category in $train$ set to replace the mean classification score in LOCE, and its results are shown in Table~\ref{table-component} (last row). We can see that the dataset-guided LOCE achieves a lower performance than the score-guided one~(\eg, 23.4 AP \vs27.4 AP for object detection). Specifically, the dataset-guided LOCE obtains a higher AP for rare classes, but it greatly hurts the performance of frequent classes. These results show that the dataset-guided method under-represents the head classes, as discussed in Section~\ref{introduction}. In contrast, the proposed score-guided method can adaptively adjust the loss margin and the sampling rate for each class according to the learning status, improving the performance for the classes from all groups. 

\vspace{-3mm}
\paragraph{Hyper-parameters.} We compare the performance with different smoothing coefficients $\alpha$ for updating the mean classification score. From the experiment results shown in Table~\ref{table-smoothing}, we observe that the performance is insensitive to the value of $\alpha$, and $\alpha=0.9$ yields the best performance. The performance with different $\hat{s}_{C+1}$ is shown in Table~\ref{table-bgscore}. We can see that the detector achieves the best performance when $\hat{s}_{C+1}=0.01$. We then conduct several experiments to study the robustness with respect to $k$ and $m$ of feature sampler in Table~\ref{table-sampler}. Through a coarse search, we set $k=8$ and $m=4$ for the rest of the experiments.

\vspace{-3mm}
\paragraph{Analysis of classification equilibrium.} Here we analyze the mean classification score and the classification accuracy between different methods on LVIS $val$ set. As shown in Figure~\ref{table-analysis}, the distribution of the mean classification score predicted by the detector trained with softmax cross-entropy loss and random image sampling is severely skewed. Specifically, the mean classification score of tail classes is close to 0, and its classification accuracy is also close to 0. When using RFS instead of random image sampling to train the detector, both the mean classification score and the classification accuracy are improved marginally. 
In contrast, the mean classification score predicted by the proposed LOCE is more balanced than those predicted by the two methods mentioned above, and the classification accuracy of common classes and tail classes is improved.

\subsection{Comparison with the State-of-the-Art}
We compare LOCE with the state-of-the-art methods on LVIS v0.5 and LVIS v1.0 in Table~\ref{table-comparison}. On LVIS v0.5, the proposed method achieves the detection performance of 28.2 AP and segmentation performance of 28.4 AP, surpassing the most recent long-tailed object detectors such as BAGS~\cite{li2020overcoming}~(by 2.4 AP and 2.1 AP) and BALMS~\cite{ren2020balanced}~(by 0.6 AP and 1.4 AP). Specifically, it outperforms BAGS by 4.0 points on rare classes, which shows the superior performance of the proposed method for tail classes. Compared with most existing methods such as~\cite{gupta2019lvis,li2020overcoming,ren2020balanced,tan2020equalization,wu2020forest}, the proposed method gets a much higher performance on head classes, in addition to improving the detection performance on tail classes.
On LVIS v1.0, the proposed method achieves a better result than all the methods shown in Table~\ref{table-comparison}, including the concurrent work such as Seesaw Loss~\cite{wang2021seesaw} and EQL v2~\cite{tan2021equalizationv2}. With the framework of Mask R-CNN~\cite{he2017mask}, LOCE achieves the detection performance of 27.4 AP and 29.0 AP on R-50-FPN and R-101-FPN, outperforming recent works such as BAGS~\cite{li2020overcoming}, Seesaw Loss~\cite{wang2021seesaw} and EQL v2~\cite{tan2021equalizationv2} by more than 1 AP. 

\section{Conclusion}
In this paper, we explore the classification equilibrium in long-tailed object detection. We propose to use the mean classification score to indicate the learning status of the model for each category, and design an equilibrium loss and a memory-augmented feature sampling method to balance the classification.
Extensive experiments show the superiority of the proposed method, which sets a new state-of-the-art in long-tailed object detection.

{\small
\bibliographystyle{ieee_fullname}
\bibliography{egbib}

\begin{thebibliography}{10}\itemsep=-1pt

\bibitem{cao2019learning}
Kaidi Cao, Colin Wei, Adrien Gaidon, Nikos Arechiga, and Tengyu Ma.
\newblock Learning imbalanced datasets with label-distribution-aware margin
  loss.
\newblock {\em arXiv preprint arXiv:1906.07413}, 2019.

\bibitem{chen2019mmdetection}
Kai Chen, Jiaqi Wang, Jiangmiao Pang, Yuhang Cao, Yu Xiong, Xiaoxiao Li,
  Shuyang Sun, Wansen Feng, Ziwei Liu, Jiarui Xu, et~al.
\newblock Mmdetection: Open mmlab detection toolbox and benchmark.
\newblock {\em arXiv preprint arXiv:1906.07155}, 2019.

\bibitem{chen2020improved}
Xinlei Chen, Haoqi Fan, Ross Girshick, and Kaiming He.
\newblock Improved baselines with momentum contrastive learning.
\newblock {\em arXiv preprint arXiv:2003.04297}, 2020.

\bibitem{deng2009imagenet}
Jia Deng, Wei Dong, Richard Socher, Li-Jia Li, Kai Li, and Li Fei-Fei.
\newblock Imagenet: A large-scale hierarchical image database.
\newblock In {\em Proceedings of the IEEE Conference on Computer Vision and
  Pattern Recognition}, pages 248--255, 2009.

\bibitem{everingham2010pascal}
Mark Everingham, Luc Van~Gool, Christopher~KI Williams, John Winn, and Andrew
  Zisserman.
\newblock The pascal visual object classes (voc) challenge.
\newblock {\em International Journal of Computer Vision}, 88(2):303--338, 2010.

\bibitem{gupta2019lvis}
Agrim Gupta, Piotr Dollar, and Ross Girshick.
\newblock Lvis: A dataset for large vocabulary instance segmentation.
\newblock In {\em Proceedings of the IEEE Conference on Computer Vision and
  Pattern Recognition}, pages 5356--5364, 2019.

\bibitem{he2020momentum}
Kaiming He, Haoqi Fan, Yuxin Wu, Saining Xie, and Ross Girshick.
\newblock Momentum contrast for unsupervised visual representation learning.
\newblock In {\em Proceedings of the IEEE Conference on Computer Vision and
  Pattern Recognition}, pages 9729--9738, 2020.

\bibitem{he2017mask}
Kaiming He, Georgia Gkioxari, Piotr Doll{\'a}r, and Ross Girshick.
\newblock Mask r-cnn.
\newblock In {\em Proceedings of the IEEE International Conference on Computer
  Vision}, pages 2961--2969, 2017.

\bibitem{he2016deep}
Kaiming He, Xiangyu Zhang, Shaoqing Ren, and Jian Sun.
\newblock Deep residual learning for image recognition.
\newblock In {\em Proceedings of the IEEE Conference on Computer Vision and
  Pattern Recognition}, pages 770--778, 2016.

\bibitem{kang2019decoupling}
Bingyi Kang, Saining Xie, Marcus Rohrbach, Zhicheng Yan, Albert Gordo, Jiashi
  Feng, and Yannis Kalantidis.
\newblock Decoupling representation and classifier for long-tailed recognition.
\newblock {\em arXiv preprint arXiv:1910.09217}, 2019.

\bibitem{li2019memory}
Suichan Li, Dapeng Chen, Bin Liu, Nenghai Yu, and Rui Zhao.
\newblock Memory-based neighbourhood embedding for visual recognition.
\newblock In {\em Proceedings of the IEEE International Conference on Computer
  Vision}, pages 6102--6111, 2019.

\bibitem{li2020overcoming}
Yu Li, Tao Wang, Bingyi Kang, Sheng Tang, Chunfeng Wang, Jintao Li, and Jiashi
  Feng.
\newblock Overcoming classifier imbalance for long-tail object detection with
  balanced group softmax.
\newblock In {\em Proceedings of the IEEE Conference on Computer Vision and
  Pattern Recognition}, pages 10991--11000, 2020.

\bibitem{lin2017feature}
Tsung-Yi Lin, Piotr Doll{\'a}r, Ross Girshick, Kaiming He, Bharath Hariharan,
  and Serge Belongie.
\newblock Feature pyramid networks for object detection.
\newblock In {\em Proceedings of the IEEE Conference on Computer Vision and
  Pattern Recognition}, pages 2117--2125, 2017.

\bibitem{lin2017focal}
Tsung-Yi Lin, Priya Goyal, Ross Girshick, Kaiming He, and Piotr Doll{\'a}r.
\newblock Focal loss for dense object detection.
\newblock In {\em Proceedings of the IEEE International Conference on Computer
  Vision}, pages 2980--2988, 2017.

\bibitem{lin2014microsoft}
Tsung-Yi Lin, Michael Maire, Serge Belongie, James Hays, Pietro Perona, Deva
  Ramanan, Piotr Doll{\'a}r, and C~Lawrence Zitnick.
\newblock Microsoft coco: Common objects in context.
\newblock In {\em Proceedings of the European Conference on Computer Vision},
  pages 740--755. Springer, 2014.

\bibitem{menon2020long}
Aditya~Krishna Menon, Sadeep Jayasumana, Ankit~Singh Rawat, Himanshu Jain,
  Andreas Veit, and Sanjiv Kumar.
\newblock Long-tail learning via logit adjustment.
\newblock {\em arXiv preprint arXiv:2007.07314}, 2020.

\bibitem{morik1999combining}
Katharina Morik, Peter Brockhausen, and Thorsten Joachims.
\newblock Combining statistical learning with a knowledge-based approach: a
  case study in intensive care monitoring.
\newblock Technical report, Technical Report, 1999.

\bibitem{ren2020balanced}
Jiawei Ren, Cunjun Yu, Xiao Ma, Haiyu Zhao, Shuai Yi, et~al.
\newblock Balanced meta-softmax for long-tailed visual recognition.
\newblock {\em Advances in Neural Information Processing Systems}, 33, 2020.

\bibitem{ren2015faster}
Shaoqing Ren, Kaiming He, Ross Girshick, and Jian Sun.
\newblock Faster r-cnn: Towards real-time object detection with region proposal
  networks.
\newblock {\em arXiv preprint arXiv:1506.01497}, 2015.

\bibitem{tan2021equalizationv2}
Jingru Tan, Xin Lu, Gang Zhang, Changqing Yin, and Quanquan Li.
\newblock Equalization loss v2: A new gradient balance approach for long-tailed
  object detection.
\newblock In {\em Proceedings of the IEEE Conference on Computer Vision and
  Pattern Recognition}, pages 1685--1694, 2021.

\bibitem{tan2020equalization}
Jingru Tan, Changbao Wang, Buyu Li, Quanquan Li, Wanli Ouyang, Changqing Yin,
  and Junjie Yan.
\newblock Equalization loss for long-tailed object recognition.
\newblock In {\em Proceedings of the IEEE Conference on Computer Vision and
  Pattern Recognition}, pages 11662--11671, 2020.

\bibitem{tian2019fcos}
Zhi Tian, Chunhua Shen, Hao Chen, and Tong He.
\newblock Fcos: Fully convolutional one-stage object detection.
\newblock In {\em Proceedings of the IEEE International Conference on Computer
  Vision}, pages 9627--9636, 2019.

\bibitem{wang2021seesaw}
Jiaqi Wang, Wenwei Zhang, Yuhang Zang, Yuhang Cao, Jiangmiao Pang, Tao Gong,
  Kai Chen, Ziwei Liu, Chen~Change Loy, and Dahua Lin.
\newblock Seesaw loss for long-tailed instance segmentation.
\newblock In {\em Proceedings of the IEEE Conference on Computer Vision and
  Pattern Recognition}, pages 9695--9704, 2021.

\bibitem{wang2020devil}
Tao Wang, Yu Li, Bingyi Kang, Junnan Li, Junhao Liew, Sheng Tang, Steven Hoi,
  and Jiashi Feng.
\newblock The devil is in classification: A simple framework for long-tail
  instance segmentation.
\newblock In {\em Proceedings of the European Conference on Computer Vision},
  pages 728--744. Springer, 2020.

\bibitem{wang2020cross}
Xun Wang, Haozhi Zhang, Weilin Huang, and Matthew~R Scott.
\newblock Cross-batch memory for embedding learning.
\newblock In {\em Proceedings of the IEEE Conference on Computer Vision and
  Pattern Recognition}, pages 6388--6397, 2020.

\bibitem{wu2020forest}
Jialian Wu, Liangchen Song, Tiancai Wang, Qian Zhang, and Junsong Yuan.
\newblock Forest r-cnn: Large-vocabulary long-tailed object detection and
  instance segmentation.
\newblock In {\em Proceedings of the 28th ACM International Conference on
  Multimedia}, pages 1570--1578, 2020.

\bibitem{xie1989logit}
Yu Xie and Charles~F Manski.
\newblock The logit model and response-based samples.
\newblock {\em Sociological Methods \& Research}, 17(3):283--302, 1989.

\bibitem{zhong2020representation}
Yujie Zhong, Zelu Deng, Sheng Guo, Matthew~R Scott, and Weilin Huang.
\newblock Representation sharing for fast object detector search and beyond.
\newblock In {\em Proceedings of the European Conference on Computer Vision},
  pages 471--487, 2020.

\end{thebibliography}
}

\clearpage


\section*{Supplementary material}

# – Supplementary material –
# Exploring Classification Equilibrium in Long-Tailed Object Detection

## 1. Implementation details

Here we elaborate on the more detailed implementations for the experiments in ablation study.

**Feature memory module.** The memory size $M$ is set to 80, which can store five latest instance features for each category. Thus, we initialize the memory features by randomly selecting 5 instances per class. In addition, an image may contain multiple instances of the same category (*e.g.*, $> 20$ instances) on LVIS. For simplicity and efficiency, we randomly select 5 instances to update the feature memory for each category that contains more than 5 instances in an image during the training.

**Dataset-based EBL.** For dataset-based EBL, we use the number of the training instances of each category to compute the loss margin as follow:

$$\delta_{yy'} = log(\frac{n_{y'}}{n_y}), \quad (1)$$

where $n_y$ is the number of the training instances of category $y$. As the score-based EBL, we use a small value $\hat{n}_{C+1}$ to replace $n_{C+1}$ (*i.e.*, the number of the training instances of background) to reduce the false positive cases. In our experiments, $\hat{n}_{C+1}$ is set to 1.

**Dataset-based MFS.** The bounding box generator and the feature memory module of dataset-based MFS are the same as those of score-based MFS, while the probabilistic sampler of dataset-based MFS uses the sampling probability based on the number of the training instances of each category. Namely, the sampling probability is computed as follow:

$$p_y = \frac{f(n_y)}{\sum_{y'} f(n_{y'})}, \quad (2)$$

where:

$$f(n_y) = \frac{1}{n_y}. \quad (3)$$

**Repeat factor sampling (RFS).** RFS over-samples the images containing the tail classes by increasing the sampling rate for these categories. As demonstrated in [1], we use the best setting of RFS that over-samples the categories that appear in less than 0.1% of the total images.

## References

[1] Agrim Gupta, Piotr Dollar, and Ross Girshick. Lvis: A dataset for large vocabulary instance segmentation. In *Proceedings of the IEEE Conference on Computer Vision and Pattern Recognition*, pages 5356–5364, 2019.



\end{document}